\documentclass[final]{cvpr}

\usepackage{times}
\usepackage{epsfig}
\usepackage{graphicx}
\usepackage{amsmath, amsbsy}
\usepackage{bm}
\usepackage{amssymb}
\usepackage{xcolor,colortbl}
\usepackage{subfig}
\usepackage[font=small]{caption}
\usepackage{pifont}
\usepackage{array}
\usepackage{bbold}
\usepackage{nicefrac}
\makeatletter
\@namedef{ver@everyshi.sty}{}
\makeatother
\usepackage{tikz}
\usepackage{tikzscale}
\usepackage{pgfplots}
\pgfplotsset{compat=newest}
\usepackage[mode=buildnew]{standalone}
\usepackage{booktabs}
\usepackage{multirow}
\usepackage{makecell}
\usepackage{siunitx}
\usepackage[accsupp]{axessibility}

\usepackage[pagebackref=true,breaklinks=true,colorlinks,bookmarks=false]{hyperref}

\def\cvprPaperID{17} 
\def\confYear{CVPR 2022}

\newdimen\owntablesep
\newcommand{\stz}{\rule{0mm}{2.3ex}}

\DeclareSymbolFont{AMSb}{U}{msb}{m}{n}
\DeclareSymbolFontAlphabet{\mathbb}{AMSb}
\DeclareMathOperator*{\argmin}{argmin}
\begin{document}

\title{On the Choice of Data\\ for Efficient Training and Validation of End-to-End Driving Models}

\author{Marvin Klingner\footnotemark[1]\quad Konstantin Müller\thanks{indicates equal contribution}\quad Mona Mirzaie \\ Jasmin Breitenstein\quad Jan-Aike Termöhlen\quad Tim Fingscheidt\\
{\tt\small \{m.klingner, konstantin.mueller, mona.mirzaie}\\[-0.2em]
{\tt\small j.breitenstein, j.termoehlen, t.fingscheidt\}@tu-bs.de}\\[0.7em]
Technische Universität Braunschweig, Braunschweig, Germany}

\maketitle

\begin{abstract}
    The emergence of data-driven machine learning (ML) has facilitated significant progress in many complicated tasks such as highly-automated driving. While much effort is put into improving the ML models and learning algorithms in such applications, little focus is put into how the training data and/or validation setting should be designed. In this paper we investigate the influence of several data design choices regarding training and validation of deep driving models trainable in an end-to-end fashion. Specifically, (i) we investigate how the amount of training data influences the final driving performance, and which performance limitations are induced through currently used mechanisms to generate training data. (ii) Further, we show by correlation analysis, which validation design enables the driving performance measured during validation to generalize well to unknown test environments. (iii) Finally, we investigate the effect of random seeding and non-determinism, giving insights which reported improvements can be deemed significant. Our evaluations using the popular \texttt{CARLA} simulator provide recommendations regarding data generation and driving route selection for an efficient future development of end-to-end driving models.
\end{abstract}
 
\section{Introduction}
\label{sec:introduction}

\begin{figure}[t]
	\centering
	\includegraphics[width=1.0\linewidth]{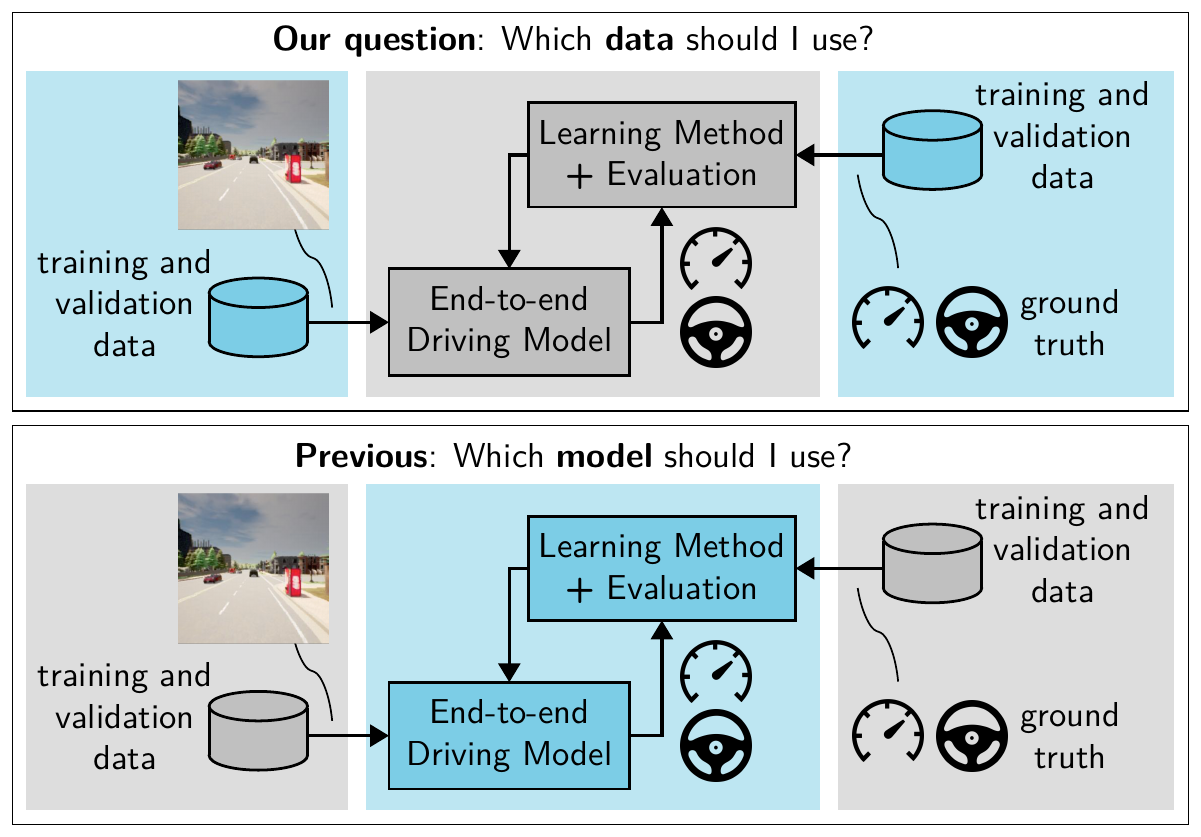}
	\vspace{-0.5cm}
	\caption{\textbf{General concept}: We investigate which data one should use to efficiently train and validate end-to-end driving models, while previous works often focused on improving models and learning methods. Blue parts in the figure identify components, optimized by many other works (bottom) and our work (top).}
	\vspace{-0.25cm}
	\label{fig:high_level}
	\vspace{-0.35cm}
\end{figure} 

\textbf{Towards End-to-End Deep Driving}:
In recent years there has been a steady trend towards higher automation levels in applications such as autonomous driving. Taking a look at the representative development in this area, past driving assistance systems have mainly been enabled by classical image processing techniques, such as using the Canny edge detector~\cite{Canny1986} with suitable post-processing to extract lanes and facilitate a lane keeping assistant~\cite{Mammeri2015}. Current systems~\cite{Badue2021} often still partially rely on such model-based algorithms, but also make use of data-driven machine learning models, e.g., for advanced environment perception tasks such as scene segmentation~\cite{Baer2020, Chen2018, Klingner2020a}, or behavior prediction of traffic participants~\cite{Jain2020, Ridel2018}. Latest developments, however, envision end-to-end trainable deep driving algorithms~\cite{Bojarski2016, Prakash2021} driven solely by high amounts of data with high-dimensional sensor measurements as input and driving signals as output as shown in Fig.~\ref{fig:high_level}. As data is one of the main factors for such methods' success, it is essential to understand the influence of different data design choices.
\par
\textbf{Training of Deep Driving Models}:
The core idea of end-to-end trainable driving algorithms~\cite{Bojarski2016} is to remove hand-crafted intermediate representations as it is nearly impossible to design a complete set of features that are important for the driving functionality. Accordingly, such end-to-end trainable models are supposed to learn an optimal representation on their own, where the learning success is highly dependent on the chosen training data. Surprisingly, despite this fact current works often rather propose alternative datasets~\cite{Codevilla2019, Dosovitskiy2017, Sun2020} instead of comparing the effect of choosing different training data. Moreover, the focus of current works is often rather on comparing different input/output representations~\cite{Chitta2021, Sauer2018}, learning methods~\cite{Codevilla2018a, Kendall2019}, or network architectures~\cite{Ishihara2021, Prakash2021}, cf.~Fig.~\ref{fig:high_level}, bottom part. In this work, we provide analysis and recommendations regarding training data and its limitations when keeping all other aspects of the deep driving model and learning method fixed, cf.~Fig.~\ref{fig:high_level}, top part. 
\par
\textbf{Evaluation of Deep Driving Models}:
Regarding evaluation of deep driving models, one can in general distinguish between open-loop evaluation (e.g., as in~\cite{Bewley2019, Hawke2020}), where the model's predictions are compared to those of an expert in an offline fashion, and closed-loop evaluation (e.g., as in~\cite{Dosovitskiy2017, Codevilla2019, Prakash2021, Chitta2021}) where the driving policy is deployed and its driving quality is measured. While Codevilla~et~al.~\cite{Codevilla2018} have shown that offline metrics correlate badly to driving quality, it is still rarely investigated, which kind of closed-loop evaluation setting is well-suited to measure driving quality. In this work, we investigate a large variety of closed-loop evaluation settings using the \texttt{CARLA} simulator~\cite{Dosovitskiy2017}. As a result, we provide guidance regarding a suitable validation design for end-to-end deep driving such that the validation performance generalizes well to the test performance, and a well-performing training checkpoint can be chosen. Furthermore, the \texttt{CARLA} simulator is subject to significant non-determinism such that evaluations cannot be carried out in a deterministic fashion. This actually reflects a real experimental setting quite well, where the same evaluation can also never be deterministically executed twice. As this phenomenon influences essentially all current works in the field, we investigate this effect to derive insights as to when a reported improvement is actually meaningful.
\par
\textbf{Contributions}: To sum up, our contributions include the following. Firstly, we investigate the effect of varying amounts of training data on the final driving performance of end-to-end deep driving models. Secondly, we provide an analysis w.r.t the limitations of currently used training data in the domain of end-to-end deep driving. Thirdly, by correlation analysis, we provide recommendations for a well-generalizing validation of driving models. Finally, we investigate random seed dependency and non-determinism in current end-to-end deep driving models, which provides insights into the meaningfulness and comparability of reported improvements in end-to-end deep driving.

\section{Related Work}
\label{sec:related_work}
In this section we discuss related work on training and evaluation of end-to-end deep driving models.
\par 
\textbf{Training of End-to-End Deep Driving Models}:
The initial work of Bojarski~et~al.~\cite{Bojarski2016} facilitated much research in end-to-end deep driving. As their simple imitation learning approach could not handle challenging urban driving scenarios, subsequent works proposed several improvements: 
Firstly, more advanced learning methods were used. Some important examples are the incorporation of navigational commands by conditional imitation learning \cite{Codevilla2018a}, the application of reinforcement learning techniques \cite{Kendall2019, Toromanoff2020, Chen2021a}, and a two-stage training, where first a privileged expert is trained whose knowledge is subsequently transferred to the driving agent \cite{Chen2020a, Zhang2021}. Constrained highway scenarios were even approached by inverse reinforcement learning \cite{Rosbach2019, Sharifzadeh2016}, where the reward is not manually designed but optimized. 
Secondly, improved network architectures making use of, e.g., long short-term memory units~\cite{Xu2017} or self-attention~\cite{Ishihara2021} have been presented. Furthermore, multi-task networks with auxiliary tasks \cite{Wang2019h, Yang2018e}, in particular with semantic segmentation \cite{Carton2021} benefit the end-to-end driving task. The fusion of different input modalities such as camera and LiDAR has also been proposed~\cite{Prakash2021}.
Thirdly, different input and output representations have been proposed. Cai~et~al.~\cite{Cai2021} show that a driving model can also be trained based on LiDAR data, while other approaches replace the direct steering and speed output by affordances~\cite{Sauer2018} or waypoints~\cite{Chitta2021} and a subsequent PID controller, or even a probabilistic output~\cite{Amini2019}.
Fourthly, the transfer of deep driving models to real data has been investigated as many deep driving models are trained and validated on simulated data and generalize poorly to real data~\cite{Osinski2020}. For example GAN-based style transfer of real images to the virtual domain~\cite{Mueller2018} or the segmentation domain~\cite{Yang2018d} has been proposed.
Finally, some works aim at improved interpretability of deep driving models by using the attention mechanism~\cite{Wei2021, Cultrera2020} or an intermediate semantic representation~\cite{Sadat2020}. 
\par 
While many aspects influencing the training of end-to-end deep driving have been thoroughly investigated, the influence of using different training sets has not been subject to a structured investigation so far, which we address with this work. Approaches introducing new datasets~\cite{Codevilla2019, Dosovitskiy2017, Sun2020, Xu2017} usually only compare different models on new data but do not show the influence induced by different amounts of training data. Other works investigate the usage of online data selection techniques~\cite{Prakash2020, Das2021} or the adaptation to out-of-distribution scenes~\cite{Filos2020}. However, these works are limited in their improvement as more online adaptation also involves catastrophic forgetting~\cite{Klingner2022} such that our recommendations for training set design are a vital component for well-performing driving policies. Moreover, we show that one of the main limitations of currently trained driving models is not the learning approach but the training data generated by a far-from-optimal ``expert'' driving policy. 
\par 
\textbf{Evaluation of End-to-End Deep Driving Models}: 
Approaches to end-to-end deep driving usually train their models using recorded driving sequences with corresponding driving actions from a human driver or an expert driving policy. Datasets such as BDD~\cite{Xu2017} or Waymo~\cite{Sun2020} collected in real environments often already provide a wide variety of situations. However, evaluation of models in a test car is usually not possible and neural simulators of ``real'' data~\cite{Kim2021} still lack performance and diversity, such that only the deviation between the predicted and the ground truth action can be measured at each time step~\cite{Bewley2019, Hawke2020}. Notably, Codevilla~et~al.~\cite{Codevilla2018} show that such offline metrics correlate badly to actual driving quality. 
In consequence, current research focus has shifted to virtual data~\cite{Richter2016, Dosovitskiy2017}, in particular to the \texttt{CARLA} simulator~\cite{Dosovitskiy2017} providing flexible possibilities to test driving models in interaction with complex and configurable environments. Accordingly, the majority of current approaches report their performance on the \texttt{CARLA} benchmarks CoRL2017~\cite{Dosovitskiy2017}, NoCrash~\cite{Codevilla2019}, and the newly introduced Leaderboard~\cite{Leaderboard}, providing possibilities to upload an agent policy for evaluation on unknown test sequences. 
While many works present new datasets or benchmarks, we provide recommendations for an efficient validation design such that the measured performance generalizes well to an unknown driving test. As the driving performance usually varies strongly during different training epochs, a good validation design also enables the selection of a well-performing model checkpoint which easily improves the model's final driving quality.

\section{End-to-End Deep Driving}
\label{sec:end_to_end_deep_driving}

In the following, we introduce our investigated problem setting of end-to-end deep driving as well as the \texttt{TransFuser} method \cite{Prakash2021}, used to approach this problem. 

\subsection{Problem Definition}
\label{sec:problem_definition}

\begin{figure}[t]
	\centering
	\includegraphics[width=1.0\linewidth]{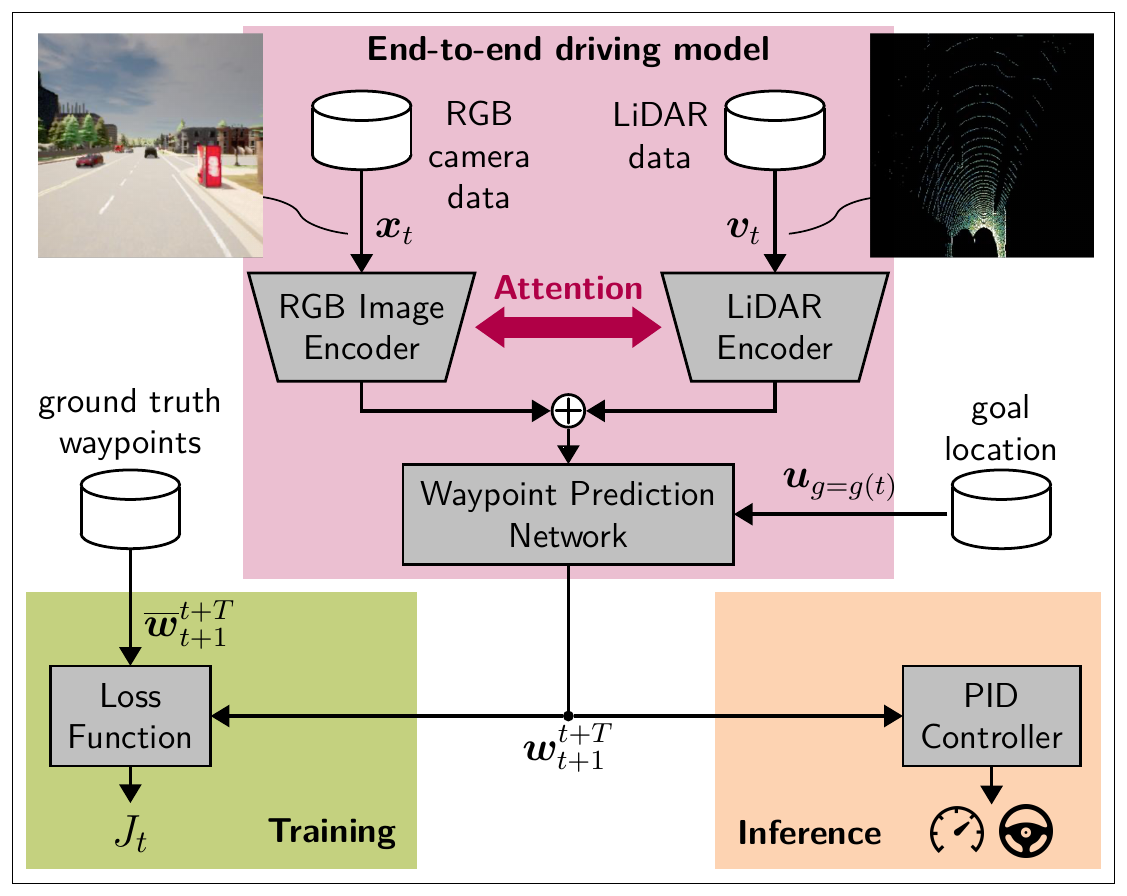}
	\vspace{-0.5cm}
	\caption{\textbf{End-to-end driving method}: The \texttt{TransFuser} model predicts steering, throttle, and brake signals (bottom right) from a camera image and a LiDAR bird's eye view image (top left and top right). Both inputs are encoded, the extracted features are fused, and finally waypoints are predicted from a GRU-based network, making use of the goal location. During training, a loss is applied minimizing the difference between predicted and ground truth waypoints (bottom left). During inference, the waypoints are converted to control signals via a PID controller (bottom right).}
	\vspace{-0.4cm}
	\label{fig:model}
\end{figure} 

\textbf{Task Description}: 
We investigate the task of point-to-point navigation in an urban environment. The goal is to drive from a starting point $\bm{u}_1\in\mathbb{R}^2$ along a pre-defined route $\bm{u}_1^G = \left(\bm{u}_1, ..., \bm{u}_g, ..., \bm{u}_G\right)$, defined by $G$ 2D waypoints $\bm{u}_g\in\mathbb{R}^2$ towards an end point $\bm{u}_G$, while following traffic rules and avoiding hazardous incidents in the interaction with other traffic participants. These sparse locations are given by the route definition as global GPS coordinates, as is standard for the \texttt{CARLA} Leaderboard. We therefore employ this approach also in our used \texttt{CARLA v0.9.13}.
\par
\textbf{Input and Output Representation}:
At each discrete time instant $t$, the model has access to several input signals provided by sensor measurements $\mathcal{X}_t$. 
Firstly, an RGB front camera image $\bm{x}_t\in\mathbb{I}^{H\times W\times C}$ with height $H=256$, width $W=256$, number of channels $C=3$, and $\mathbb{I}=\left[0,1\right]$ is available. Note that the camera images are extracted at a resolution of $300\times 400$ pixels, which we crop to the region of interest at resolution of $256\times 256$, thereby also removing artifacts at the edges.
Secondly, a LiDAR point cloud converted to a histogram pseudo-image $\bm{v}_t\in\mathbb{I}^{H\times W\times C'}$ with $C'=2$ channels in bird's eye view (BEV) is available. To generate the histogram pseudo-image, the point cloud is divided along the ground plane such that the two channels represent the histogram over the number of points at each image location on/below and over the ground plane, respectively. The underlying LiDAR point cloud is considered in a region of $\SI{32}{m}$ in front of the vehicle and $\SI{16}{m}$ to each side. From the BEV perspective, the $\SI{32}{m}\times \SI{32}{m}$ region is divided into $256\times 256$ blocks of equal size. 
Finally, the next goal waypoint $\bm{u}_{g=g(t)}$ can be used as additional input, where $g(t)$ yields the index $g$ which is the desired next goal waypoint at time instance $t$. Accordingly, the model input is defined by $\mathcal{X}_t = \left\lbrace \bm{x}_t, \bm{v}_t, \bm{u}_{g=g(t)}\right\rbrace$.
\par
As output, the series of the next $T$ waypoints $\bm{w}^{t+T}_{t+1} = \left(\bm{w}_{t+1}, ..., \bm{w}_{t+\tau}, ..., \bm{w}_{t+T}\right)$ with $\bm{w}_{t+\tau}\in\mathbb{R}^2$ of the car's future trajectory in BEV space shall be predicted. The current position and orientation serve as reference coordinate frame to the waypoint's coordinates. The predicted waypoints are converted to steering, throttle, and brake signals via a PID controller~\cite{Chen2020a}, expecting $T=4$ waypoints by default. Note that the model output, i.e., the series of waypoints $\bm{w}^{t+T}_{t+1}$, is in close proximity to the ego-vehicle, while the inputted goal waypoints $\bm{u}_{1}^{G}$ of the desired route are usually quite sparse and often further away from each other.

\subsection{Method Description}
\label{sec:method_description}

\textbf{End-to-end Driving Model}: 
We choose the \texttt{TransFuser} architecture~\cite{Prakash2021} depicted in Fig.~\ref{fig:model} for our experiments as it is one of the current state-of-the-art models for end-to-end deep driving. The architecture processes both camera image $\bm{x}_t$ and LiDAR BEV pseudo-image $\bm{v}_t$ by modality-specific \texttt{ResNet} encoders~\cite{He2016}. At each intermediate feature resolution global context information is exchanged between both encoders via attention modules~\cite{Vaswani2017a}. Thereby, both images are compressed into 512-dimensional feature vectors, which are added element-wise and passed to the waypoint prediction network, cf.~Fig.~\ref{fig:model}. This network first reduces the feature dimensionality from 512 to 64 by fully connected layers. Afterwards the network uses a GRU-based layer~\cite{Cho2014a} taking the goal location $\bm{u}_{g=g(t)}$ as additional input and a subsequent linear layer to predict the differences $\Delta \bm{w}_\tau$ between two future waypoints such that $\bm{w}_{t+\tau} = \bm{w}_{t+\tau - 1} + \Delta \bm{w}_\tau$. Note that the GRU-based layer uses its recurrent nature to predict each difference $\Delta \bm{w}_\tau$ by a separate forward pass from $\bm{w}_{t+\tau - 1}$ (using $\bm{w}_{t}=(0,0)$ for the first forward pass). The hidden state used in the first GRU layer forward pass is initialized by the previously extracted 64-dimensional feature vector. Subsequent GRU layer forward passes take the previous one's hidden state (=output) as initial hidden state. For additional details we refer to~\cite{Prakash2021}.
\par
\textbf{Training by Conditional Imitation Learning}:
Following many recent works~\cite{Chitta2021, Codevilla2018, Prakash2021}, we employ conditional imitation learning (CIL), where we aim at obtaining a driving policy $\bm{\pi}$ that is trained in a supervised fashion to imitate the driving behavior of an expert policy $\overline{\bm{\pi}}$. The driving policy $\bm{\pi}$ takes the sensor measurements $\mathcal{X}_t$ as input and outputs the future waypoint trajectory $\bm{w}^{t+T}_{t+1}$ such that
\begin{equation}
    \bm{w}^{t+T}_{t+1} = \bm{\pi}\left(\mathcal{X}_t\right).
\end{equation}
For a certain time instance $t$, also the expert driving policy can be rolled out in the environment using the same initial conditions as for the trainable driving agent. Then, the expert's driving decisions can be obtained which we represent by a series of ground truth waypoints $\overline{\bm{w}}^{t+T}_{t+1} = \left(\overline{\bm{w}}_{t+1}, ..., \overline{\bm{w}}_{t+\tau}, ..., \overline{\bm{w}}_{t+T}\right)$, $\overline{\bm{w}}_{t+\tau}\in\mathbb{R}^2$ in BEV space. To optimize the driving model, we minimize the distance between the driving model's output $\bm{w}^{t+T}_{t+1}=\bm{\pi}\left(\mathcal{X}_t\right)$ and the expert policy's output $\overline{\bm{w}}^{t+T}_{t+1}$ using the mean absolute error
\begin{equation}
    J_t = J\left(\bm{\pi}\left(\mathcal{X}_t\right), \overline{\bm{w}}^{t+T}_{t+1}\right)=\sum_{\tau=1}^T|| \bm{w}_{t+\tau} - \overline{\bm{w}}_{t+\tau}||_1 
\end{equation}
as shown in the bottom left of Fig.~\ref{fig:model}. If we now reinterpret $t$ as a sample index such that we consider a whole dataset $\mathcal{D} = \cup_{r=1}^R \mathcal{D}_r$ consisting of $R$ routes $\mathcal{D}_r = \left\lbrace (\mathcal{X}_t, \overline{\bm{w}}^{t+T}_{t+1}), t\in\left\lbrace1,...,N_r\right\rbrace\right\rbrace$ of (possibly varying) length $N_r$ with sensor measurements and corresponding expert driving decisions, we can optimize the model as 
\begin{equation}
    \bm{\pi}^* = \argmin_{\bm{\pi}} \mathbb{E}_{(\mathcal{X}_t,\overline{\bm{w}}^{t+T}_{t+1})\sim\mathcal{D}}       \left[J\left(\bm{\pi}\left(\mathcal{X}_t\right), \overline{\bm{w}}^{t+T}_{t+1}\right)\right],
\end{equation}
to obtain the optimal driving policy $\bm{\pi}^*$. In practice, we implement the driving model using \texttt{PyTorch}~\cite{Paszke2019} and train it for 50 epochs using the AdamW optimizer~\cite{Loshchilov2019} with a learning rate of $10^{-4}$ and weight decay of $0.01$. 
\par
\textbf{Inference using a PID Controller}:
While the model is trained to predict waypoints in BEV space, the final driving actions are determined by an inverse dynamics model~\cite{Chen2020a} implemented as PID controller, cf.~bottom right in Fig.~\ref{fig:model}. Specifically, there are two separate PID controllers for both lateral and longitudinal control, both taking the driving model's predicted future waypoints $\bm{w}^{t+T}_{t+1}$ as input. Accordingly, the longitudinal controller sets throttle and brake, while the lateral controller sets the steering. Specific implementation details can be found in~\cite{Prakash2021}.

\section{Training and Evaluation Setup}
\label{sec:training_evaluation}

We conduct our experiments using the latest \texttt{CARLA v0.9.13}. In the following, we describe our data generation process as well as our validation and test design.

\subsection{Training Dataset Generation}
\label{sec:train_set_generation}

\begin{figure}[t]
	\centering
	\includegraphics[width=1.0\linewidth]{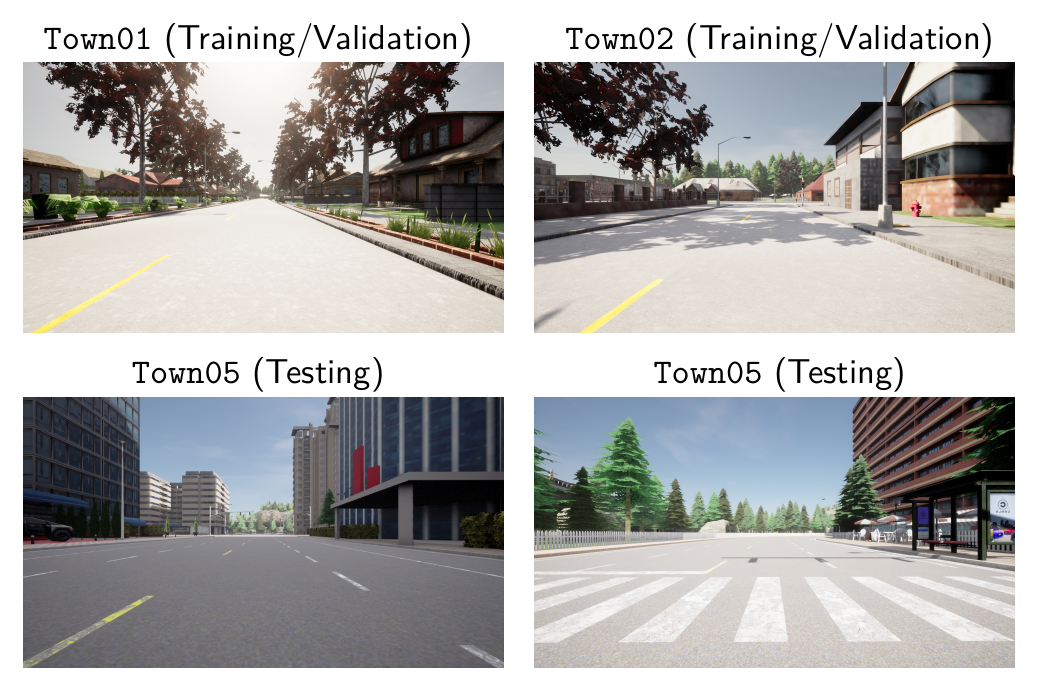}
	\vspace{-0.6cm}
	\caption{\textbf{\texttt{CARLA} simulation environment}: We show some exemplary images, collected in towns, used for training and validation (top) and from the town used for testing (bottom).}
	\vspace{-0.2cm}
	\label{fig:data_samples}
\end{figure} 

\textbf{Data Generation Concept}:
For training data, we follow the protocol of \cite{Prakash2021} and roll out an expert policy in \texttt{CARLA}, recording the observations $\mathcal{X}_t$ and corresponding expert actions $\overline{\bm{w}}^{t+T}_{t+1}$ at a frame rate of \SI{2}{fps}. RGB images are captured with a forward-facing camera at a resolution of $400 \times 300$ pixels and field of view (FOV) of $100^{\circ}$. LiDAR point clouds are captured with a ray-cast-based \texttt{Velodyne 64} LiDAR at a rotation frequency of \SI{10}{fps} at $10^{\circ}$ upper FOV and $-30^{\circ}$ lower FOV. Further, the handcrafted expert policy used to generate $\overline{\bm{w}}^{t+T}_{t+1}$ has access to privileged simulator information to avoid collisions and other infractions.
\par 

\begin{table}[t]
  \centering
  \caption{\textbf{Training set design}: The datasets we use mainly differ in the number of collected images and the number of used routes. We also report the corresponding portions of the four most frequent driving maneuvers (last four columns) given in (\%). All training data $\mathcal{D}^{\mathrm{train}}$ has been collected in \texttt{CARLA Town[01-04, 06-07, 10]}, while \texttt{CARLA Town05} is kept for testing.}
  \vspace{-0.2cm}
  \resizebox{\columnwidth}{!}{
  \setlength{\tabcolsep}{6pt}
  \begin{tabular}{c|cc|cccc}
  training & \multirow{2}{*}{\# images} & \multirow{2}{*}{\# routes} & follow  & go & turn & turn \\
  set &  &  & lane & straight & left & right \\
  \hline\hline
  $\mathcal{D}^{\mathrm{train}}_{\mathrm{100K}}$\stz & 99,806 & 1762 & 69.8 & 11.3 & 6.9 & 10.3 \\
  $\mathcal{D}^{\mathrm{train}}_{\mathrm{160K}}$\stz & 166,852 & 1903 & 71.5 & 10.2 & 6.9 & 9.5 \\
  $\mathcal{D}^{\mathrm{train}}_{\mathrm{220K}}$\stz & 228,023 & 2901 & 69.5 & 11.4 & 8.2 & 9.1 \\
  \end{tabular}}
  \vspace{-0.25cm}
  \label{tab:train_set_properties}
\end{table}

\textbf{Route Design Considerations}:
The expert follows a set of predefined routes (uniquely defined by sparse waypoints $\bm{u}_1^G$), during which the expert encounters several complex urban traffic scenarios. We collect training data in seven different \texttt{CARLA} towns (cf.~Tab.~\ref{tab:train_set_properties}) ranging from rural areas, residential districts to urban areas under simple \texttt{ClearNoon} weather conditions (cf.~Fig.~\ref{fig:data_samples}), as we do not focus on investigations regarding weather domain shift. Along a route, the expert is exposed to various pre-defined randomly picked traffic scenarios, even some scenarios where other traffic participants do not adhere to traffic rules. The expert navigates along two different route types (cf.~Fig.~\ref{fig:route_design}) in each town: \textit{Tiny routes} (T) involve a single traffic intersection or turn and are usually shorter than 100 meters. \textit{Short routes} (S) cover two or more intersections, being typically 300 to 500 meters long. The training dataset does not include \textit{long routes} (L), which involve complex routing with a total length of more than 1000 meters, as these are characterized by a large imbalance of driving maneuvers towards ``follow lane''. Accordingly, for each town a set of tiny and short routes is generated. As we investigate different amounts of training data, we ensured that the distribution of driving maneuvers is approximately the same for all collected datasets, cf.~Tab.~\ref{tab:train_set_properties}. Still, some more variety regarding traffic scenarios is to be expected in larger datasets, which can hardly be quantified.

\subsection{Validation and Test Design}
\label{sec:val_test_setup}

\begin{figure}[t]
	\centering
	\subfloat[Long routes (L)\label{fig:long_routes}]{\includegraphics[width=0.14\textwidth]{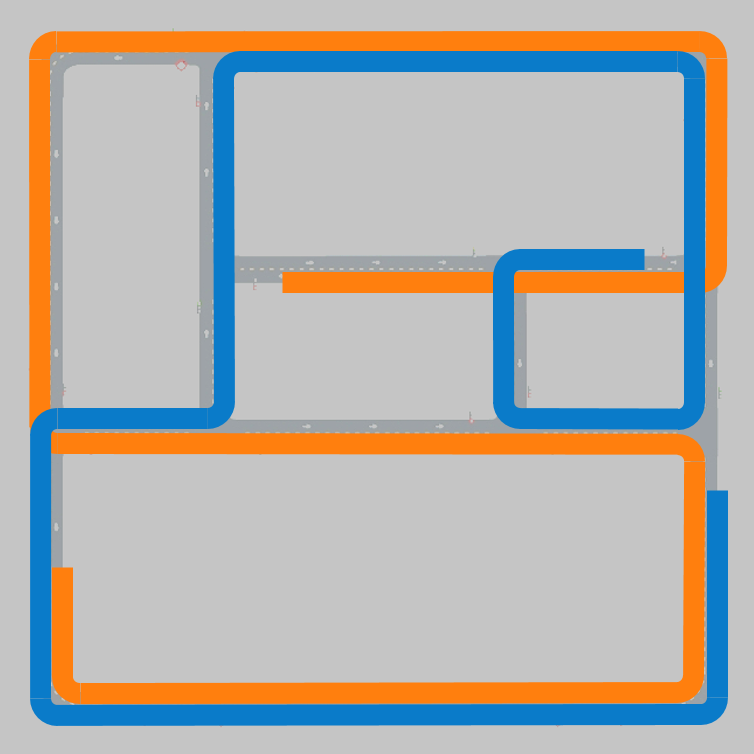}}\hfill
	\subfloat[Short routes (S)\label{fig:short_routes}]{\includegraphics[width=0.14\textwidth]{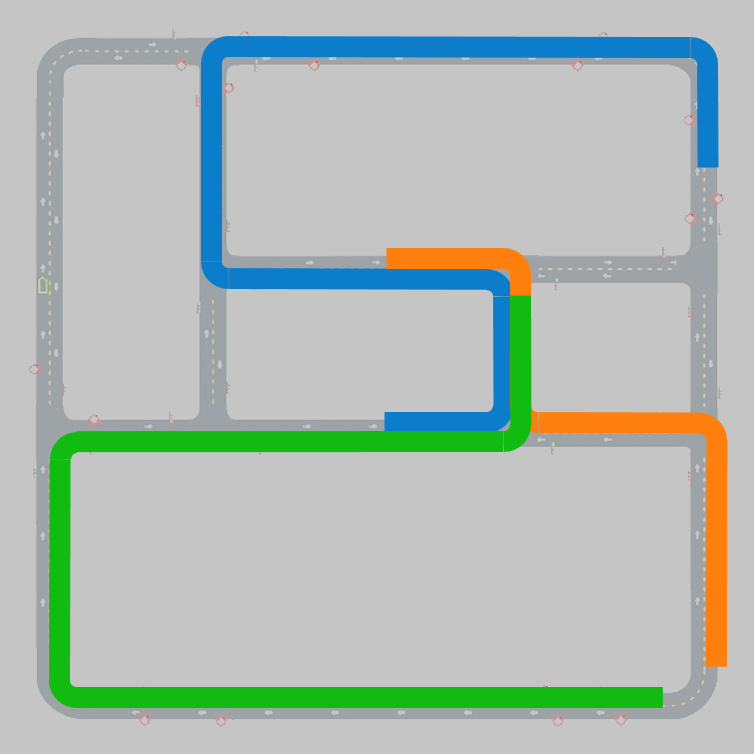}}\hfill
	\subfloat[Tiny routes (T)\label{fig:tiny_routes}]{\includegraphics[width=0.14\textwidth]{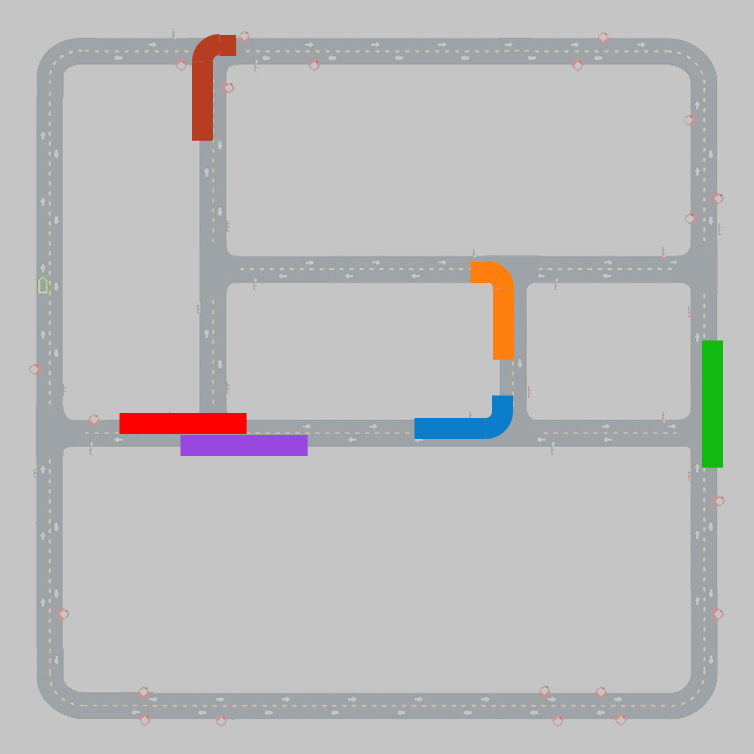}}
	\caption{\textbf{Validation and test route types}: We show examples of the different route types which the driving agent should drive along during validation and testing. Long routes (L) usually progress over many intersections and turns, while short routes (S) only cover a few of these urban traffic sections. A tiny route (T), on the other hand, only covers a single such section by design.}
	\vspace{-0.2cm}
	\label{fig:route_design}
\end{figure} 

\textbf{Validation and Test Design}:
For validation and testing, the driving quality of the trained model is measured when driving along pre-defined routes. Note that such closed-loop evaluation is only possible in simulation and differs significantly from many fields, where only the model predictions on a single-image basis are evaluated in open-loop fashion. We put emphasis on comparing the effect of using long, short, or tiny routes depicted in Figs.~\ref{fig:long_routes}-\ref{fig:tiny_routes}, respectively, for validation. We aim at a validation whose performance generalizes well to the test performance. For testing, we employ 10 long routes from \texttt{CARLA Town05} as in~\cite{Prakash2021}. Note that \texttt{CARLA Town05} has not been seen during training and validation.
\par

\begin{table}[t]
  \centering
  \caption{\textbf{Validation and test design}: The validation and test setups mainly differ in the number and length of their routes. We also report the corresponding portions of the four most frequent driving maneuvers (last four columns) given in (\%). Moreover, test routes $\mathcal{R}^{\mathrm{test}}$ are located in \texttt{CARLA Town05}, while validation routes $\mathcal{R}^{\mathrm{val}}$ are located in \texttt{CARLA Town[01-04, 06-07]}.}
  \vspace{-0.1cm}
  \resizebox{\columnwidth}{!}{
  \setlength{\tabcolsep}{5pt}
  \begin{tabular}{c|cc|cccc}
  evaluation & \multirow{2}{*}{\# routes} & route & follow & go & turn & turn \\
  routes & & type & lane & straight & left & right\\
  \hline\hline
  $\mathcal{R}^{\mathrm{val}}_{\mathrm{160T}}$\stz & 160 & Tiny &   45.3 & 19.2 & 17.7 & 16.7 \\
  $\mathcal{R}^{\mathrm{val}}_{\mathrm{80T}}$\stz & 80 & Tiny &   45.1 & 15.2 & 20.9 & 17.5 \\
  $\mathcal{R}^{\mathrm{val}}_{\mathrm{22S,1}}$\stz & 22 & Short & 75.4 & 7.7 & 9.4 & 6.9 \\
  $\mathcal{R}^{\mathrm{val}}_{\mathrm{22S,2}}$\stz & 22 & Short & 72.4 & 10.7 & 9.1 & 7.3 \\
  $\mathcal{R}^{\mathrm{val}}_{\mathrm{11S}}$\stz & 11 & Short & 80.1 & 4.5 & 8.7 & 6.2 \\
  $\mathcal{R}^{\mathrm{val}}_{\mathrm{12L}}$\stz & 12 & Long & 78.3 & 10.3 & 4.2 & 6.9 \\
  $\mathcal{R}^{\mathrm{val}}_{\mathrm{6L}}$\stz & 6 & Long & 77.5 & 10.3 & 3.5 & 8.4 \\
  \hline
  $\mathcal{R}^{\mathrm{test}}$\stz & 10 & Long & 77.9 & 11.2 & 4.9 & 5.1 \\
  \end{tabular}}
  \vspace{-0.2cm}
  \label{tab:val_set_properties}
\end{table}

\textbf{Validation Route Generation}:
For validation, we generate new routes as reported in Tab.~\ref{tab:val_set_properties} based on a method proposed by \cite{Prakash2021}. First, intersections are located on the map of a \texttt{CARLA} town based on the position of traffic lights. Then for each route a start waypoint $\bm{u}_1$ and an end waypoint $\bm{u}_G$ is sampled from the vertices of a square of size $\SI{100}{m} \times \SI{100}{m}$ centered around an identified intersection. Based on these two
waypoints, a trajectory is generated by a global route planning algorithm as in \cite{Leaderboard} forming a sparse sequence of route waypoints $\bm{u}_1^G$. This technique typically produces short or long routes with two or more intersections. Each passing of an intersection or turn in these routes can be converted into a tiny route by selecting the start point before the respective location and the end point afterwards. A more detailed description of the route generation process is given in \cite{Prakash2021}. After route generation, we remove duplicates in the generated routes as the used algorithm does not ensure a unique route design.

\section{Experiments}
\label{sec:experiments}

In the following, we explain our evaluation methodology for analyzing data design choices. Afterwards, we investigate validation route and training data design.

\begin{figure}[t]
	\centering
	\includegraphics[width=1.0\linewidth]{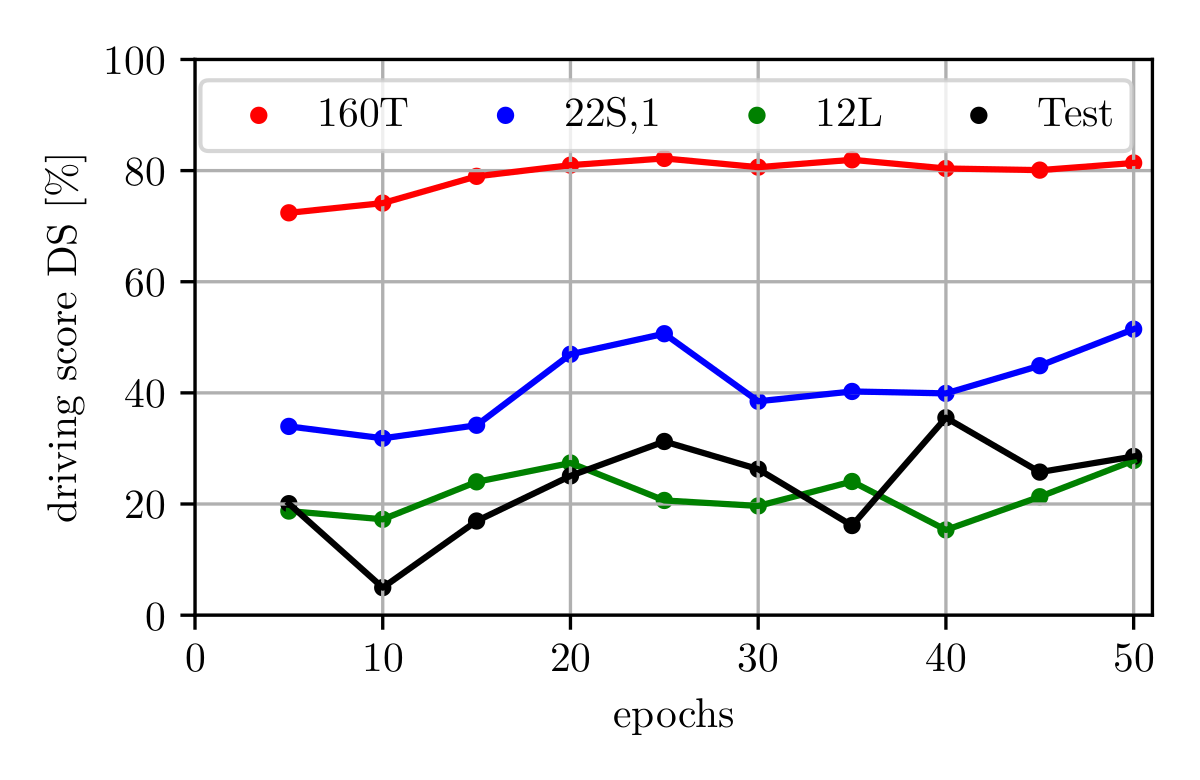}
	\vspace{-0.8cm}
	\caption{\textbf{Driving scores} $\mathrm{DS}$ measured at different epochs on \textbf{different validation sets} $\mathcal{R}^{\mathrm{val}}_{(\cdot)}$, e.g., 12L represents $\mathcal{R}^{\mathrm{val}}_{\mathrm{12L}}$, and on the test set $\mathcal{R}^{\mathrm{test}}$ (identified by ``Test''). Training is performed on $\mathcal{D}^{\mathrm{train}}_{160K}$. We observe high variance in performance over the course of training, making it difficult to identify the best driving model.}
	\vspace{-0.2cm}
	\label{fig:val_set_scores}
\end{figure} 

\subsection{Evaluation Methodology}
\label{sec:eval_metrics}

\textbf{Driving Performance}: 
We follow recent works~\cite{Chen2021a, Chitta2021, Prakash2021, Toromanoff2020} in using the driving score $\mathrm{DS}\in\left[0,1\right]$ as the main metric to measure the driving performance of a model. As outlined in Sec.~\ref{sec:val_test_setup}, the validation of driving models is conducted on a set of pre-defined routes. The driving score considers two aspects regarding driving quality along these routes: First, the route completion percentage $\mathrm{RC}\in\left[0,1\right]$ is calculated. Possible error cases lowering $\mathrm{RC}$ are, deviations from the pre-defined route (i.e., route deviation), an agent not taking any decisions (i.e., blocked agent), a route not finished in time (i.e., route timeout), or off-road driving. Second, an infraction score $\mathrm{IS}\in\left[0,1\right]$ considering various traffic incidents is computed as defined in~\cite{Leaderboard}. Considered infractions lowering $\mathrm{IS}$ are collisions with pedestrians, other vehicles, and static elements, as well as running red lights or stop signs. For our investigations regarding a suitable validation design, we report the driving score as it reflects the overall driving quality, which we aim at optimizing. For an in-depth analysis regarding limitations of currently used training data for driving models, we additionally report statistics on all considered error cases. 
\par
\textbf{Correlation and Generalization}: 
During the course of training, the driving score obtained on the test set may vary significantly (black line in Fig.~\ref{fig:val_set_scores}). As a consequence, we would desire our validation to reflect these performance changes well. As the optimal test driving score is usually reached before the final training epoch 50, we train our driving models for 50 epochs, validate and test it every 5 epochs, and compute the Pearson and/or Spearman correlation between the obtained driving scores. Moreover, we compute the driving score on the test set using the optimal model selected during validation to investigate the generalizability of our validation to the test set. Note that we use this methodology being aware of the test set performance only to find out, which validation design has a good predictive power for the performance obtained during testing.

\subsection{Validation Route Design}
\label{sec:validation_set_design}
 
\textbf{Varying Driving Performance During Training}:
Starting point for our data design investigations was the experiment shown in Fig.~\ref{fig:val_set_scores}. The driving score $\mathrm{DS}$ obtained by the driving model on the test routes $\mathcal{R}^{\mathrm{test}}$ (black line) varies strongly between $5\,...\,35$ for different training epochs. Even towards the end of training there is no stable convergence, which was a typical behavior in all of our experiments. \textit{Our conclusion is that a good checkpoint selection is essential}, as the checkpoint after 50 epochs often turned out to perform poorly (cf.~first row in Tab.~\ref{tab:val_sets_scores}). 
\par 

\begin{figure}[t]
	\centering
	\includegraphics[width=1.0\linewidth]{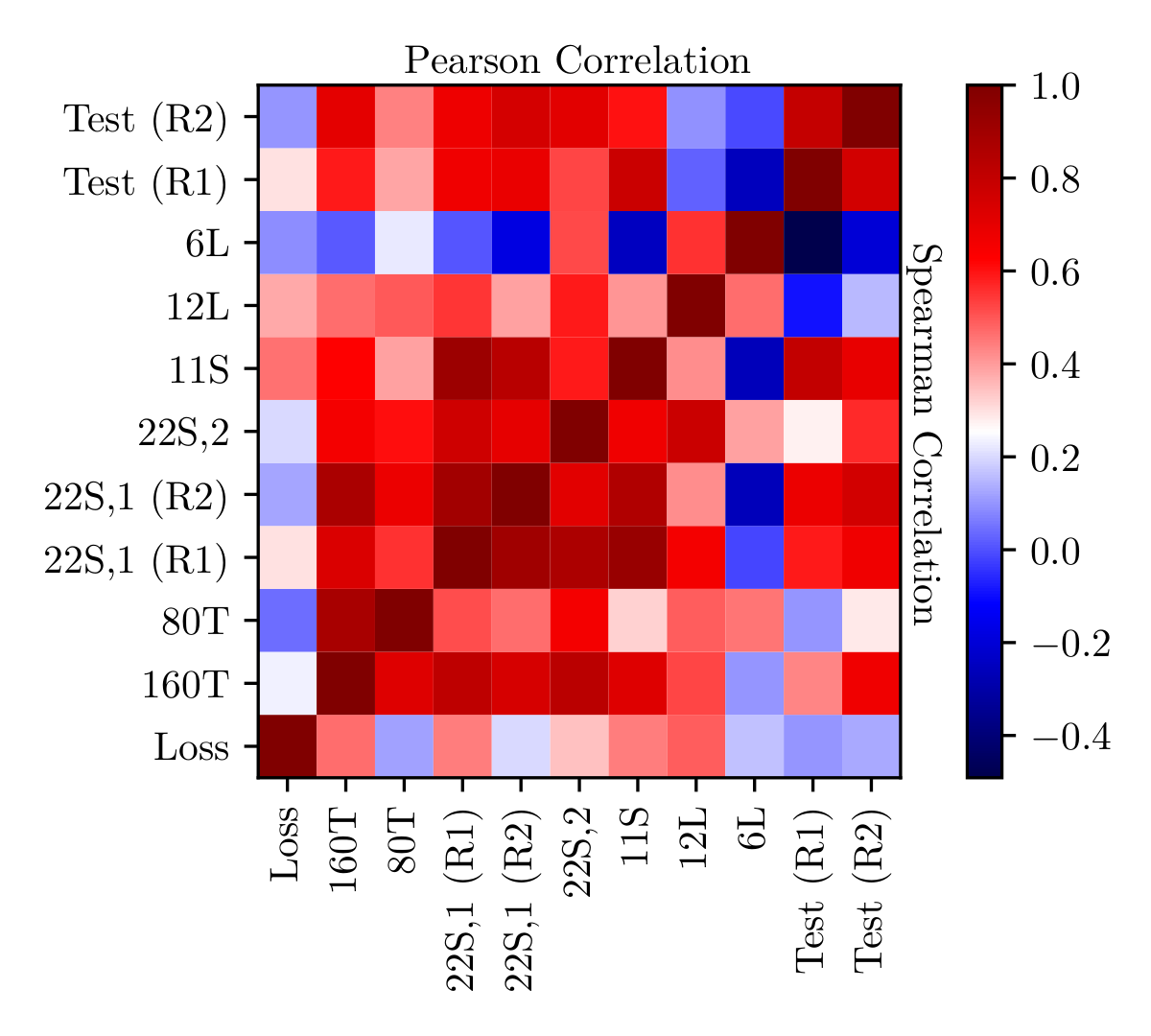}
	\vspace{-0.6cm}
	\caption{Pearson correlation (upper left part) and Spearman correlation (lower right part) between the driving scores measured on various validation routes $\mathcal{R}^{\mathrm{val}}_{(\cdot)}$, e.g., 12L represents $\mathcal{R}^{\mathrm{val}}_{\mathrm{12L}}$, and the test routes $\mathcal{R}^{\mathrm{test}}$ (identified by ``Test''). R1 and R2 represent two different random seeds. We also report correlations between the validation loss ``Loss'' obtained as in~\cite{Prakash2021} and the driving scores. }
	\label{fig:val_set_corr}
\end{figure} 

\textbf{Validation Performance Correlation}:
After this initial observation, our goal was to find a good generalizing validation, as the test set performance is usually unknown. To get an initial overview, we investigated the correlation between the performance measured on several differently designed validation routes (inducing similar computation complexity) and test routes for a single training run in Fig.~\ref{fig:val_set_corr}. Several interesting insights are observable in this figure: First, the offline validation loss (computed as in~\cite{Prakash2021}) correlates badly with driving performance $\mathrm{DS}$ on all validation and test routes, which is expected, considering the results for offline metrics from Codevilla~et~al.~\cite{Codevilla2018}. Second, the driving performance $\mathrm{DS}$ on test routes ($\mathcal{R}^{\mathrm{test}}$) consisting of 10 long routes correlates rather poorly to the validation performance $\mathrm{DS}$ on 12 or 6 other long routes ($\mathcal{R}^{\mathrm{val}}_{\mathrm{12L}}$ and $\mathcal{R}^{\mathrm{val}}_{\mathrm{6L}}$). As a reason we observed that for long routes, single events such as a blocked agent have a comparably large influence on the overall driving score as only few routes are considered. Moreover, such events happen with different frequency on different routes due to different difficulty level. Even on the same route due to the non-determinism of the validation and test simulations in \texttt{CARLA}, long routes are particularly volatile in their performance variations across different validations as can be seen from the Test or 12L curve in Fig.~\ref{fig:val_set_scores}. Short routes ($\mathcal{R}^{\mathrm{val}}_{\mathrm{22S,1}}$) and in particular tiny routes ($\mathcal{R}^{\mathrm{val}}_{\mathrm{160T}}$), on the other hand, are usually (a bit) less volatile due to the averaging over more routes. Third, in this initial experiment we observe a high correlation between the validation performance on a medium sized set of short routes ($\mathcal{R}^{\mathrm{val}}_{\mathrm{22S,1}}$, $\mathcal{R}^{\mathrm{val}}_{\mathrm{22S,2}}$, and $\mathcal{R}^{\mathrm{val}}_{\mathrm{11S}}$) and the test routes or the set containing 12 long routes ($\mathcal{R}^{\mathrm{val}}_{\mathrm{12L}}$). Similar observations can be made for tiny routes (160T and 80T), although they correlate a bit worse when looking at the Spearman correlation. The performance on the set of 6 long routes ($\mathcal{R}^{\mathrm{val}}_{\mathrm{6L}}$) has no high correlation with any other validation performance, again underlining the high performance variability of long routes. 
\par 

\begin{table}[t]
  \centering
  \caption{\textbf{Pearson correlation} between the \textbf{validation set performance} and the \textbf{test set performance} (given by the driving score). We show results for different models trained on $\mathcal{D}^{\mathrm{train}}_{(\cdot)}$ (cf.~Tab.~\ref{tab:train_set_properties}) and for different validation sets, i.e., all models were validated on $\mathcal{R}^{\mathrm{val}}_{(\cdot)}$ (cf.~Tab.~\ref{tab:val_set_properties}), where R1 and R2 represent two different random seeds. Best results in boldface, second-best underlined.}
  \vspace{-0.15cm}
  \resizebox{\columnwidth}{!}{
  \setlength{\tabcolsep}{2pt}
  \begin{tabular}{cl|c|cccc}
   & & time per & \multicolumn{4}{c}{trained on}\\
   & & checkpoint & $\mathcal{D}^{\mathrm{train}}_{\mathrm{100K}}$\stz & $\mathcal{D}^{\mathrm{train}}_{\mathrm{160K}}$ (R1) & $\mathcal{D}^{\mathrm{train}}_{\mathrm{160K}}$ (R2) & $\mathcal{D}^{\mathrm{train}}_{\mathrm{220K}}$\\
  \hline\hline
  \multirow{8}{*}{\rotatebox{90}{validated on}} &  $\mathcal{R}^{\mathrm{val}}_{\mathrm{160T}}$\stz & $\sim\SI{5}{h}$                                                        & \underline{0.26} & 0.68 & 0.30 & -0.02 \\
  & $\mathcal{R}^{\mathrm{val}}_{\mathrm{80T}}$ & $\sim\SI{2.5}{h}$       & 0.13 & 0.43 & 0.40 & -0.01 \\
  & $\mathcal{R}^{\mathrm{val}}_{\mathrm{22S},1}$ (R1) & $\sim\SI{5}{h}$  & 0.22 & 0.71 & 0.73 & \textbf{0.65} \\
  & $\mathcal{R}^{\mathrm{val}}_{\mathrm{22S},1}$ (R2) & $\sim\SI{5}{h}$  & 0.01 & \textbf{0.76} & 0.73 & \underline{0.53} \\
  & $\mathcal{R}^{\mathrm{val}}_{\mathrm{22S},2}$ & $\sim\SI{5}{h}$       & 0.15 & 0.65 & \textbf{0.82} & 0.47 \\
  & $\mathcal{R}^{\mathrm{val}}_{\mathrm{11S}}$ & $\sim\SI{2.5}{h}$       & 0.03 & \underline{0.73} & 0.76 & 0.51 \\
  & $\mathcal{R}^{\mathrm{val}}_{\mathrm{12L}}$ & $\sim\SI{5}{h}$         & \textbf{0.79} & 0.06 & \underline{0.79} & 0.39 \\
  & $\mathcal{R}^{\mathrm{val}}_{\mathrm{6L}}$ & $\sim\SI{2.5}{h}$        & 0.22 & -0.14 & 0.70 & 0.33 \\
  \end{tabular}}
  \vspace{-0.25cm}
  \label{tab:val_sets_corrs}
\end{table}

\textbf{Which validation routes should be used?}
To get more conclusive evidence, which validation route design provides the best predictive power towards test set performance $\mathrm{DS}$, we compare the correlation of the driving model's performance $\mathrm{DS}$ on different validation routes for models trained on different training data in Tab.~\ref{tab:val_sets_corrs}. We choose three sets of routes, i.e.,  $\mathcal{R}^{\mathrm{val}}_{\mathrm{160T}}$, $\mathcal{R}^{\mathrm{val}}_{\mathrm{22S},1}$, and $\mathcal{R}^{\mathrm{val}}_{\mathrm{12L}}$, inducing similar computational complexity of 5 hours validation time per checkpoint. We observe that tiny routes $\mathcal{R}^{\mathrm{val}}_{\mathrm{160T}}$ lead to a rather poor correlation to the test performance (cf.~Tab.~\ref{tab:val_sets_corrs}, first row). Comparing with Fig.~\ref{fig:val_set_scores}, we suspect that tiny routes are less informative for measuring the driving quality as only a single traffic section needs to be solved per route, which is often quite easy such that the driving score is consistently high in this validation setting. In contrast, driving requires good driving behavior across many sections, which is better captured by short or long routes shown by the higher correlation values when using $\mathcal{R}^{\mathrm{val}}_{\mathrm{22S},1}$ or $\mathcal{R}^{\mathrm{val}}_{\mathrm{12L}}$ for validation. Halving the validation time per epoch by using just half the amount of routes only works reasonably well for short routes, cf. results for $\mathcal{R}^{\mathrm{val}}_{\mathrm{11S}}$ in Tab.~\ref{tab:val_sets_corrs}. 

\begin{table}[t]
  \centering
  \caption{\textbf{Test driving scores} given in (\%) obtained on $\mathcal{R}^{\mathrm{test}}$, \textbf{having used different validation sets}. We show results for four different models trained on $\mathcal{D}^{\mathrm{train}}_{(\cdot)}$ (cf.~Tab.~\ref{tab:train_set_properties}). The model checkpoint used to obtain the driving score has been selected using the validation set $\mathcal{R}^{\mathrm{val}}_{(\cdot)}$ (cf.~Tab.~\ref{tab:val_set_properties}) of the respective row. R1 and R2 represent two different random seeds. We additionally show results when testing the model after 50 epochs of training (``naive approach''), using the model obtaining the lowest validation loss (``validation loss'') and the result of the expert driving policy, which can be interpreted as an upper performance bound (``expert perf.''). Best results in boldface, second-best underlined.} 
  \vspace{-0.15cm}
  \resizebox{\columnwidth}{!}{
  \setlength{\tabcolsep}{5pt}
  \begin{tabular}{cl|cccc}
   & & \multicolumn{4}{c}{trained on}\\
   & & $\mathcal{D}^{\mathrm{train}}_{\mathrm{100K}}$\stz & $\mathcal{D}^{\mathrm{train}}_{\mathrm{160K}}$ (R1) & $\mathcal{D}^{\mathrm{train}}_{\mathrm{160K}}$ (R2) & $\mathcal{D}^{\mathrm{train}}_{\mathrm{220K}}$\\
  \hline\hline
  & naive approach & 8.9 & \underline{26.3} & 17.5 & 15.8 \\
  & validation loss & \underline{13.8} & 16.9 & 19.8 & 19.6 \\
  \hline
  \multirow{8}{*}{\rotatebox{90}{validated on}} &  $\mathcal{R}^{\mathrm{val}}_{\mathrm{160T}}$\stz & 13.4 & \textbf{32.7} & 13.7 & 26.3 \\
  & $\mathcal{R}^{\mathrm{val}}_{\mathrm{80T}}$         & 13.4 & 17.7 & 26.5 & 15.8 \\
  & $\mathcal{R}^{\mathrm{val}}_{\mathrm{22S},1}$ (R1)  & 12.6 & \underline{26.3} & 26.5 & \underline{26.7} \\
  & $\mathcal{R}^{\mathrm{val}}_{\mathrm{22S},1}$ (R2)  & 12.6 & \underline{26.3} & \underline{28.4} & 20.5 \\
  & $\mathcal{R}^{\mathrm{val}}_{\mathrm{22S},2}$       & 12.6 & \textbf{32.7} & \underline{28.4} & 26.3 \\
  & $\mathcal{R}^{\mathrm{val}}_{\mathrm{11S}}$         & 12.6 & \underline{26.3} & 26.5 & 26.3 \\
  & $\mathcal{R}^{\mathrm{val}}_{\mathrm{12L}}$         & \textbf{22.3} & \underline{26.3} & \textbf{28.9} & \textbf{30.8} \\
  & $\mathcal{R}^{\mathrm{val}}_{\mathrm{6L}}$\stz      & 12.6 & 17.7 & 16.1 & 15.8 \\
  \hline
  & expert perf.                                        & 46.8 & 46.8 & 46.8 & 46.8 \\
  \end{tabular}}
  \vspace{-0.4cm}
  \label{tab:val_sets_scores}
\end{table}
\begin{table*}[t]
  \centering
  \caption{\textbf{Performance} on \textbf{different training sets}: We report driving score $\mathrm{DS}$, route completion $\mathrm{RC}$, and infraction score $\mathrm{IS}$ for various training sets containing a different number of samples. Values are given in (\%) and higher is better. We also show the test performance of the expert (``expert perf.''), used during generation of training data $\mathcal{D}^{\mathrm{train}}$. Results are averaged over three test runs. As we observed quite some differences between different test runs, we also report respective standard deviations. We additionally report metrics regarding route completion failures and infraction types in number of events per kilometer where lower is better.}
  \vspace{-0.2cm}
  \resizebox{\linewidth}{!}{
  \setlength{\tabcolsep}{3pt}
  \begin{tabular}{c|c|c|c|cccc|c|ccccc}
  training & time per& driving & route & route & agent & route & off-road & infraction & collisions with & collisions with & collisions with & running a & running a \\
  set & epoch & score & completion & deviation & blocked & timeout & driving & score & pedestrians & other vehicles & static elements & red light & stop sign\\
  \hline\hline
  $\mathcal{D}^{\mathrm{train}}_{\mathrm{100K}}$ & $\sim\SI{0.5}{h}$ & $16.8\!\pm\! 4.6$ & $91.0\!\pm\! 5.3$ & $0.0\!\pm\! 0.0$ & $2.4\!\pm\! 1.9$ & $0.5\!\pm\! 0.4$ & $3.3\!\pm\! 0.1$ & $18\!\pm\! 6$ & $3.6\!\pm\! 0.7$ & $9.6\!\pm\! 0.3$ & $0.3\!\pm\! 0.5$ & $40.2\!\pm\! 3.8$ & $5.6\!\pm\! 2.6$ \\
  $\mathcal{D}^{\mathrm{train}}_{\mathrm{160K}}$ & $\sim\SI{1.0}{h}$ & $28.8\!\pm\! 2.6$ & $80.7\!\pm\! 3.3$ & $0.0\!\pm\! 0.0$ & $6.7\!\pm\! 2.1$ & $0.3\!\pm\! 0.5$ & $2.1\!\pm\! 1.0$ & $37\!\pm\! 2$ & $2.3\!\pm\! 0.5$ & $6.3\!\pm\! 1.6$ & $1.2\!\pm\! 1.0$ & $24.6\!\pm\! 1.4$ & $3.5\!\pm\! 3.0$ \\
  $\mathcal{D}^{\mathrm{train}}_{\mathrm{220K}}$ & $\sim\SI{2.0}{h}$ & $32.2\!\pm\! 4.8$ & $78.6\!\pm\! 4.8$ & $0.0\!\pm\! 0.0$ & $11.8\!\pm\! 6.9$ & $0.0\!\pm\! 0.0$ & $2.8\!\pm\! 0.9$ & $44\!\pm\! 2$ & $1.0\!\pm\! 0.1$ & $5.5\!\pm\! 1.8$ & $1.5\!\pm\! 2.6$ & $21.6\!\pm\! 1.8$ & $4.2\!\pm\! 2.3$ \\
  \hline
  expert perf. & - & $46.8\!\pm\! 7.2$ & $83.4\!\pm\! 4.2$ & $0.0\!\pm\! 0.0$ & $5.6\!\pm\! 2.9$ & $0.0\!\pm\! 0.0$ & $0.0\!\pm\! 0.0$ & $60\!\pm\! 8$ & $1.0\!\pm\! 0.9$ & $8.2\!\pm\! 2.9$ & $0.0\!\pm\! 0.0$ & $8.6\!\pm\! 3.2$ & $0.0\!\pm\! 0.0$ \\
  \end{tabular}}
  \vspace{-0.2cm}
  \label{tab:data_amount}
\end{table*}

\par 
Looking at the obtained driving scores in Tab.~\ref{tab:val_sets_scores}, this fact is confirmed when validating the model every 5 epochs and choosing the best-performing checkpoint for testing. First of all, we observe that the performance of the naive approach in Tab.~\ref{tab:val_sets_scores}, which simply tests the driving model obtained after 50 epochs of training is often quite low. Using the validation loss for model selection improves slightly, however, independent of the used validation routes, the test performance is almost always better or on par when using the checkpoint selection. Again we observe that tiny routes $\mathcal{R}^{\mathrm{val}}_{\mathrm{160T}}$ tend to lead to poor driving scores due to the rather weak correlation leading to a bad checkpoint selection. Well performing model checkpoints are usually selected by short routes $\mathcal{R}^{\mathrm{val}}_{\mathrm{22S, 1}}$ and long routes $\mathcal{R}^{\mathrm{val}}_{\mathrm{12L}}$, with results much closer to the expert's performance. However, when halving the number of routes, the set $\mathcal{R}^{\mathrm{val}}_{\mathrm{11S}}$ still yields high test performance, while $\mathcal{R}^{\mathrm{val}}_{\mathrm{6L}}$ produces rather poor results due to the aforementioned high volatility when using few long routes. \textit{As conclusion, we would choose a large number of long routes when having access to a vast amount of computation resources. If one aims at efficient model development, a medium-sized set of short routes (such as, e.g., $\mathcal{R}^{\mathrm{val}}_{\mathrm{11S}}$) provides a good trade-off between validation time and driving performance} as shown in Tab.~\ref{tab:val_sets_scores}. Moreover, when noting that the driving performance does not improve anymore through further training, the training process can be stopped, thereby reducing training time by well-generalizing validation during training.
\par
\textbf{Effect of Non-Determinism}:
As we observed quite some variance in the obtained driving performance, we want to give insights into the effect of different validation design choices. We provide results in Tabs.~\ref{tab:val_sets_corrs} and \ref{tab:val_sets_scores}. First, we simply repeated the validation on $\mathcal{R}^{\mathrm{val}}_{\mathrm{22S, 1}}$ using a different random seed (R1,R2). The results regarding correlation and driving score are similarly high but differ in quite a bit in some cases. Similar observations are made when running the same training with a different random seed ($\mathcal{D}^{\mathrm{train}}_{\mathrm{160K}}$ (R1) and $\mathcal{D}^{\mathrm{train}}_{\mathrm{160K}}$ (R2)) or when varying the chosen validation routes ($\mathcal{R}^{\mathrm{val}}_{\mathrm{22S, 1}}$ and $\mathcal{R}^{\mathrm{val}}_{\mathrm{22S, 2}}$). However, we observed that simulations in \texttt{CARLA} currently cannot be run in completely deterministic fashion such that the same evaluation result can never be reproduced completely. This is actually similar to an experimental setting in reality, where this is also the case. Therefore, \textit{the standard deviation of results gives important insights into the reproducibility of driving models and the meaningfulness of reported results. Accordingly, it should always be reported.} For example, highly overlapping standard deviation intervals might indicate a low probability that a reported improvement is actually significant.  

\subsection{Training Data Design}
\label{sec:training_set_design}

\textbf{Which training data amount should be used?}
According to our results on a suitable validation, we now use the validation on $\mathcal{R}^{\mathrm{val}}_{\mathrm{22S, 1}}$ to select a suitable checkpoint from driving model trainings making use of different amounts of training data. After a checkpoint has been selected the reported results in Tab.~\ref{tab:data_amount} are obtained by averaging over three test set evaluations. We also report corresponding standard deviations for each result. We observe that increasing the amount of training data samples $\mathcal{X}_t$ from $100,000$ ($\mathcal{D}^{\mathrm{train}}_{\mathrm{100K}}$) to $220,000$ ($\mathcal{D}^{\mathrm{train}}_{\mathrm{220K}}$) significantly increases the driving score probably due to the larger and more diverse data basis, cf.~the respective number of routes in Tab.~\ref{tab:train_set_properties}. Interestingly, with less training data, the driving model has a very high route completion score but a rather bad infraction score. With increasing data amount the driving agent apparently learns to avoid hazardous traffic incidents (maybe there have been more diverse examples in the training data available), which, however, also makes the route completion more difficult. \textit{Training on approximately $160,000$ images already seems to provide a good trade-off between performance and complexity, while best results are obtained using larger but computationally more expensive amounts of data.}
\par
\textbf{Expert Performance Bound}:
Finally, we compare our best results to the results of the \texttt{CARLA} expert. We note that the route completion result is already similar to the expert's result for all trained models. Interestingly, even the expert gets blocked (``agent blocked'' in Tab.~\ref{tab:data_amount}) quite often. As the driving model is trained on data generated using the expert, this behavior seems to transfer to some degree. We observe a similar behavior for collision infractions. Not adhering to rules imposed by signs or traffic lights shows a slightly different behavior. The performance on these infractions tends to improve with more data but is still much worse than the expert result. Here, additional training signals might be necessary. Still, overall \textit{we conclude that for a further improvement of the driving model better data not generated by a far-from-perfect expert is essential.}

\section{Conclusions}
\label{sec:conclusion}

In this work we present recommendations regarding training and validation data for end-to-end deep driving models. Our results show that in the range of currently employed amounts of data, the driving performance still scales with more data, but seems to be strongly limited by the performance of the expert driving policy used to generate the data. Further, we find that a medium-sized set of short validation routes provides an efficient and (mostly) well-suited validation w.r.t.~generalization to unseen test data. Finally, we observe that non-determinism still influences all currently reported results on \texttt{CARLA} evaluations, showing the need to report standard deviations in reported improvements, in particular in the domain of end-to-end deep driving. We believe that our investigations will help researchers to choose efficient setups for their training data and validation design, allowing to find better models for end-to-end deep driving and to reach their goals in shorter time.

{\small
\bibliographystyle{ieee_fullname}
\bibliography{bib/ifn_spaml_bibliography} 
}

\end{document}